\documentclass[letterpaper, 10 pt, conference]{ieeeconf}  

\IEEEoverridecommandlockouts                              

\usepackage{graphics} 
\usepackage{epsfig} 
\usepackage{amsmath}
\usepackage{amssymb}  
\usepackage{bm}  
\usepackage{makecell}
\usepackage[noadjust]{cite} 
\usepackage{url}
\usepackage{hyperref}
\usepackage{booktabs} 
\usepackage{graphicx}
\usepackage{textcomp}
\usepackage{float}
\usepackage{subcaption}
\usepackage{siunitx}

\usepackage{lipsum}
\usepackage{algorithm}
\usepackage{algpseudocode}
\usepackage{setspace}
\usepackage{multirow} 
\usepackage{tabularx} 
\usepackage{array} 
\usepackage{titlesec}

\titlespacing*{\section}
{0pt}{8pt plus 2pt minus 2pt}{8pt plus 2pt minus 2pt}

\algrenewcommand\algorithmicrequire{\textbf{Input:}}
\algrenewcommand\algorithmicensure{\textbf{Output:}}

\usepackage{optidef}
\usepackage{caption}
\usepackage{subcaption}
\usepackage{tikz}
\usetikzlibrary{shapes,arrows.meta,arrows,positioning,calc}

\algnewcommand{\IRequire}{\item[\textbf{Input:}]}

\DeclareMathOperator*{\argmin}{arg\,min}

\title{\LARGE \bf
Data-Driven Sampling Based Stochastic MPC for \\
Skid-Steer Mobile Robot Navigation
}

\author{Ananya Trivedi,$^{1}$ Sarvesh Prajapati,$^{1}$ Anway Shirgaonkar,$^{1}$  Mark Zolotas,$^{1,2}$, and Ta\c{s}k{\i}n Pad{\i}r$^{1,3}$ 
\thanks{$^{1}$Institute for Experiential Robotics, Northeastern University, Boston, Massachusetts, USA. { \tt\small \{trivedi.ana, prajapati.s, shirgaonkar.a, m.zolotas, t.padir\}@northeastern.edu}}
\thanks{$^{2}$Mark Zolotas is currently at Toyota Research Institute (TRI), Cambridge, MA, USA. This paper describes work performed at Northeastern University and is not associated with TRI.}
\thanks{$^{3}$Ta\c{s}k{\i}n Pad{\i}r holds concurrent appointments as a Professor of Electrical and Computer Engineering at Northeastern University and as an Amazon Scholar. This paper describes work performed at Northeastern University and is not associated with Amazon.}
}

\begin{document}
\tikzset{
block/.style = {draw, fill=white, rectangle, minimum height=3em, minimum width=3em},
tmp/.style  = {coordinate}, 
sum/.style= {draw, fill=white, circle, node distance=1cm},
input/.style = {coordinate},
output/.style= {coordinate},
pinstyle/.style = {pin edge={to-,thin,black}
}
}
\maketitle
\thispagestyle{empty}
\pagestyle{empty}

\begin{abstract}
Traditional approaches to motion modeling for skid-steer robots struggle with capturing nonlinear tire-terrain dynamics, especially during high-speed maneuvers. In this paper, we tackle such nonlinearities by enhancing a dynamic unicycle model with Gaussian Process (GP) regression outputs. This enables us to develop an adaptive, uncertainty-informed navigation formulation. We solve the resultant stochastic optimal control problem using a chance-constrained Model Predictive Path Integral (MPPI) control method. This approach formulates both obstacle avoidance and path-following as chance constraints, accounting for residual uncertainties from the GP to ensure safety and reliability in control. Leveraging GPU acceleration, we efficiently manage the non-convex nature of the problem, ensuring real-time performance. Our approach unifies path-following and obstacle avoidance across different terrains, unlike prior works which typically focus on one or the other. We compare our GP-MPPI method against unicycle and data-driven kinematic models within the MPPI framework. In simulations, our approach shows superior tracking accuracy and obstacle avoidance. We further validate our approach through hardware experiments on a skid-steer robot platform, demonstrating its effectiveness in high-speed navigation. The GPU implementation of the proposed method and supplementary video footage are available at \url{https://stochasticmppi.github.io}.
\end{abstract}

\section{Introduction} \label{sec:introduction}

Skid-steer robots, recognized for their high traction and payload capacity, are widely adopted in rugged and challenging terrain navigation~\cite{skid_steer_survey}. However, their high maneuverability comes at the cost of significant skidding and slipping, making it challenging to predict their motion given command velocities. Accurate motion models are crucial for effective control, yet existing approaches based on simple kinematic models struggle to capture complex tire-terrain interactions~\cite{kinematic_modeling_ssmr_1}. 

\begin{figure}[t]
    \centering
    \includegraphics[width=0.83\columnwidth]{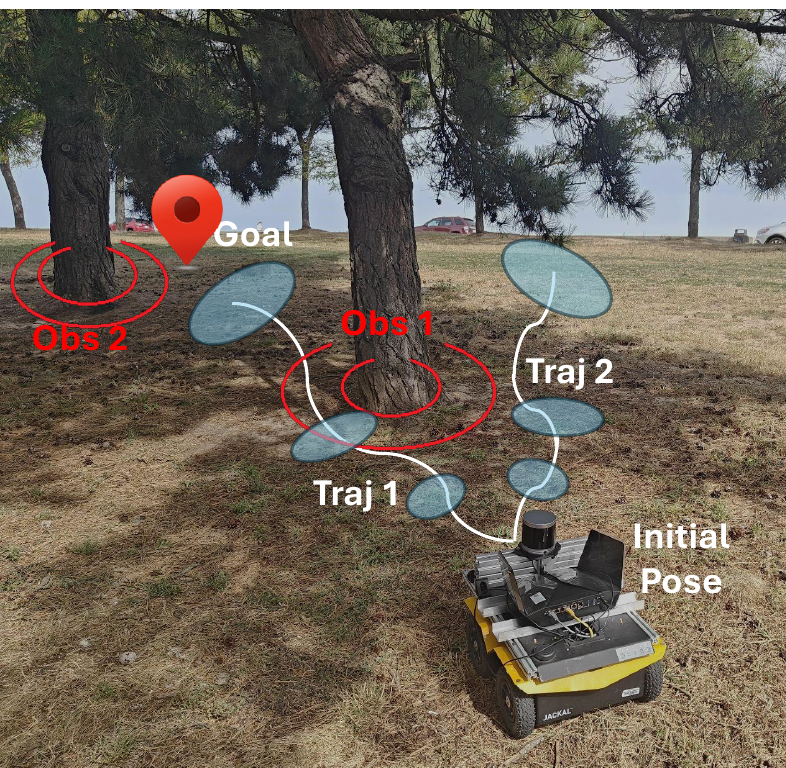} 
    \caption{Illustration of two MPPI trajectories (Traj 1 and Traj 2). The obstacle radii are adjusted based on propagated state variance (ellipses) and the safety threshold \(p_x\) (Eq.~\ref{eq:generic_chance_constraint}). Traj 1 is `unsafe' due to a collision with an obstacle (Obs 1).}
    \label{fig:approach}
    \vspace{-3.4mm}
\end{figure}

Model Predictive Path Integral (MPPI)~\cite{mppi_basic} is a sampling-based control strategy well-suited for solving non-linear, non-convex, and non-differentiable optimal control problems. It evaluates multiple candidate control trajectories and selects the optimal action based on a weighted sum of these trajectories. Nevertheless, standard MPPI approaches~\cite{mppi_different_paper} that rely on pre-trained models to predict robot dynamics struggle with unmodeled terrain variations and real-time uncertainty~\cite{chance_constrained_mppi}, limiting their ability to adapt to different environments.

To address these challenges, we integrate Gaussian Processes (GPs) into the MPPI framework~\cite{gp_textbook}. GPs are powerful nonlinear regression tools that provide both mean and uncertainty (variance) estimates for predictions. In our work, GPs model skid-steer robot motion across varying terrains. The GP mean captures model residuals, accounting for discrepancies between nominal physics-based models and actual robot dynamics. Additionally, the GP variance, representing uncertainty in the model’s predictions, is integrated into safety constraints through chance constraints~\cite{chance_constraints_pavone}. This enables safety buffers to be dynamically adjusted in real-time given varying uncertainty levels. These buffers help ensure the robot stays on its intended path and maintains a safe distance from obstacles, even in highly dynamic environments. Fig.~\ref{fig:approach} presents an overview of our approach.

The key contributions of this paper are:
\begin{itemize}
    \item  A data-driven, sampling based stochastic MPC method tailored for high-speed skid-steer robot navigation at speeds up to 2 m/s (the top speed of the Clearpath Jackal platform), leveraging GP regression for adaptive motion planning. The same control law ensures navigation across all terrains included in the GP training dataset, without the need for explicit terrain identification.
    \item A GPU implementation of the GP-MPPI planner that solves stochastic optimal control problems at real-time rates of 20 Hz for path tracking and obstacle avoidance. 
    \item Extensive simulation experiments of the proposed method, demonstrating superior tracking accuracy and obstacle avoidance compared to kinematic model-based MPPI approaches. The method's real-world applicability is also evaluated and verified via hardware experiments of a skid-steer robot navigating different terrains. 
\end{itemize}

\section{Related Work} \label{sec:related_work}
Existing skid-steer robot motion models often rely on data-driven approaches, calibrating kinematic parameters with offline data~\cite{kinematic_modeling_ssmr_1,kinematic_modeling_ssmr_2,kinematic_modeling_ssmr_3}. However, these models struggle with complex tire-terrain interactions, especially at high speeds. Dynamic models attempt to address this but studies reported in literature are typically limited to simulations and face computational inefficiencies~\cite{dynamic_modeling_ssmr_1,dynamic_modeling_ssmr_2,dynamic_modeling_ssmr_3,rabiee}. To mitigate these issues, we use Gaussian Process Regression (GPR) to capture uncertainty in skid and slip dynamics, making the system more adaptive to varying terrain conditions.

GPR has previously been utilized to model nonlinear dynamics and uncertainties in motion planning. Hewing et al.~\cite{gp_melanie} integrated GPR into an MPC framework for race cars, while Torrente et al.~\cite{gp_quadrotor} used similar methods for quadrotor trajectory tracking. While effective at motion planning, GPR is computationally expensive due to the need for GP predictions at each step. Existing methods often resort to computationally intensive solvers, like Sequential Convex Programming~\cite{nocedal_and_wright} to handle GPR within MPC. In contrast, our approach avoids solving GPR-based MPC as a single optimization problem by instead using GPU parallelization to evaluate multiple GP predictions across different candidate trajectories. This reduces computational overhead and allows MPC to be solved with real-time performance at 20 Hz. 

MPPI methods~\cite{mppi_basic} provide a natural fit for real-time adaptive control by evaluating multiple candidate control trajectories in parallel. Nevertheless, these approaches have relied on deterministic models or pre-trained neural networks~\cite{mppi_different_paper}, which can struggle with unmodeled terrain variations and uncertainties. In our approach, we establish uncertainty-aware MPPI by introducing chance constraints into the formulation given variance estimates through GPR. This approach creates a more robust, stochastic MPC solution that probabilistically satisfies safety constraints~\cite{smpc_survey}. Additionally, by computing weighted sums of the predictions from multiple terrains, we can reactively adjust to different operating conditions. 

Prior work on skid-steer robots using GPR has primarily focused on path tracking, employing methods such as robust control~\cite{angela_gp} or feedback linearization~\cite{jfr_gp}. In contrast, our approach integrates custom-designed cost functions for both path tracking and obstacle avoidance within our proposed GP-MPPI framework, providing a unified solution for navigation in complex, real-world environments.

\section{Preliminaries} \label{sec:preliminaries}
In this section, we provide an overview of the stochastic motion modeling framework introduced in our previous work~\cite{icra24,icaa}, and then detail our formulation for high-speed, uncertainty-informed navigation of skid-steer robots.

\subsection{Probabilistic Motion Modeling}\label{sec:probabilistic_motion_modeling}
The state-space of the robot includes its position, orientation, and velocities, represented by \( \boldsymbol{x} = [X, Y, \theta, v, \omega]^\top \in \mathbb{R}^5 \). Although the robot traverses rough terrain, we employ a planar model as an approximation, with terrain-induced effects captured through GPR. The control actions are the commanded linear and angular velocities, \( \boldsymbol{u} = [v_{\text{ref}}, \omega_{\text{ref}}]^\top \in \mathbb{R}^2 \). Robot positions and orientations are obtained by integrating the robot’s velocities over time. Discretizing the unicycle dynamic motion model~\cite{dynamic_unicycle}, the update equations for the robot's velocities are:
\begin{equation} \label{eq:dynamic_discrete_v}
    v(k+1) = f_v\left(v(k), \omega(k), v_{\text{ref}}(k)\right) + \overline{\delta}_v(k)
\end{equation}
\begin{equation} \label{eq:dynamic_discrete_omega}
    \omega(k+1) = f_\omega\left(v(k), \omega(k), \omega_{\text{ref}}(k)\right) + \overline{\delta}_\omega(k)
\end{equation}
Here, \( f_v \) and \( f_\omega \) are functions derived from the nominal physics-based unicycle model, describing the predicted evolution of the robot’s linear and angular velocities based on the current state and control inputs. The terms \(\overline{\delta}_v(k)\) and \(\overline{\delta}_\omega(k)\) represent unmodeled dynamics, such as cumulative nonlinear tire forces, which are challenging to model explicitly.

The key takeaway from our previous work~\cite{icra24,icaa} is that the \(\overline{\delta}_v(k)\) and \(\overline{\delta}_\omega(k)\) terms representing disturbances in linear and angular velocities are vital for accurate modeling of skid and slip. These terms are also terrain-dependent. To model such disturbances, we train two GPs per terrain—one for linear velocity and one for angular velocity. The training dataset for these GPs is constructed by computing the residual modeling error between the measured velocities and those predicted by the nominal model.

The final values of \(\overline{\delta}_v(k)\) and \(\overline{\delta}_\omega(k)\) at each time step \(k\) are a weighted sum of the different GP outputs associated with each terrain~\cite{ensemble_gp}. The weights for this summation are computed through a convex optimization problem that considers the history of robot states, comparing how the predicted values from different terrains match with what actually occurred. The weighted ensemble of GPR is calculated as:
\begin{align} \label{eq:weighted_gp_eqns}
[\overline{\delta}_v(k), \overline{\delta}_\omega(k)]^\top &\sim\mathcal{N}\left(\overline{\boldsymbol{m}}(k), \overline{\boldsymbol{C}}(k)\right) \nonumber \\ &= \mathcal{N}\left(\sum_{i=1}^M w_i \boldsymbol{m}_i(k), \sum_{i=1}^M w_i^2 \boldsymbol{C}_i(k)\right)
\end{align}
where \( M \) is the number of different terrains, \( \boldsymbol{m}_i(k) \in \mathbb{R}^2 \) and \( \boldsymbol{C}_i(k) \in \mathbb{R}^{2 \times 2} \) are the mean and covariance estimates of the GP model for terrain \( i \) at time step \( k \), and \( w_i \) is the weight associated with the GP output for terrain \( i \). We model \(M=3\) terrains in our experiments: grass, asphalt, and tile. 

To compute the optimal weights \( \boldsymbol{w}^* = [w_1^*,\ldots, w_M^*]^\top \), the following convex optimization problem is solved:
\begin{align} \label{eq:convex_opt}
    \argmin_{\boldsymbol{w}} \, & ||\boldsymbol{Y}_v-\boldsymbol{F}_v\boldsymbol{w}||_2^2 + ||\boldsymbol{Y}_\omega-\boldsymbol{F}_\omega\boldsymbol{w}||_2^2 +  \gamma||\boldsymbol{w}-\boldsymbol{w}^{*}_{k-1}||_1 \nonumber \\
    &\text{s.t.} \quad 0 \leq w_i \leq 1, \quad \sum_{i=1}^{M}w_i = 1
\end{align}
where \( \boldsymbol{Y}_v \) and \( \boldsymbol{Y}_\omega \) represent the ground truth velocities over a history of \( H \) steps, \( \boldsymbol{F}_v \) and \( \boldsymbol{F}_\omega \) are the corresponding GP mean predictions, and \( \gamma \) is a regularization parameter that penalizes deviations from the previous weight estimates \( \boldsymbol{w}^{*}_{k-1} \). Using these optimal weights, \(\boldsymbol{w}^*\), a convex combination of the outputs from all \(M\) GPs provides the best estimate for the unmodeled dynamics, \(\overline{\delta}_v(k)\) and \(\overline{\delta}_\omega(k)\), thereby improving the robot's ability to adapt to varying terrain conditions.

\subsection{Stochastic Navigation Problem}\label{sec:stochastic_navigation_problem}
We formulate the navigation task for skid-steer robots as a stochastic optimal control problem over a finite horizon $N$. Starting from an initial condition \( \boldsymbol{x}_0 \), the objective is to minimize the expected value of a cost function, $c(\cdot,\cdot)$, that depends on both the state and control inputs, subject to uncertain dynamics and probabilistic constraints: 
\begin{equation} \label{eq:cost_function}
    \min_{\boldsymbol{x}(k), \boldsymbol{u}(k)} \mathbb{E}\left[\sum_{k=0}^{N-1} c(\boldsymbol{x}(k), \boldsymbol{u}(k))\right]
\end{equation}
\begin{equation} \label{eq:dynamics}
    \text{s.t. \hspace{0.2em}} \boldsymbol{x}(k+1) = \boldsymbol{f}(\boldsymbol{x}(k), \boldsymbol{u}(k)) + \boldsymbol{\overline{\delta}}(k)
\end{equation}
\begin{equation} \label{eq:generic_chance_constraint}
    \Pr\left(\boldsymbol{x}(k) \in \mathcal{X}_{\text{safe}}\right) \geq p_x
\end{equation}
\begin{equation} \label{eq:control_constraints}
    \boldsymbol{u}(k) \in [\boldsymbol{u}_{\text{min}}, \boldsymbol{u}_{\text{max}}]
\end{equation}
\begin{equation} \label{eq:initial_condition}
    \boldsymbol{x}(0) = \boldsymbol{x}_0
\end{equation}
Here, the system's evolution is governed by the nominal dynamics $\boldsymbol{f}$ augmented by the GP correction terms $\boldsymbol{\overline{\delta}}(k) = [0, 0, 0, \overline{\delta}_v(k), \overline{\delta}_\omega(k)]^\top$, as shown in Eq.~\ref{eq:dynamics}. The safe region \( \mathcal{X}_{\text{safe}} \) includes obstacle-free areas and desired track boundaries that the robot must navigate within. Uncertainty in the robot's state stems from the Gaussian-distributed velocities modeled by the GP, which propagate through to the robot's positions. Due to the unbounded nature of these distributions, strict constraint satisfaction cannot always be guaranteed, but it can be enforced with high probability through the chance constraint parameter \( p_x \) expressed in Eq.~\ref{eq:generic_chance_constraint}. As these stochastic dynamics unfold, chance constraints ensure that safety and performance requirements are met reliably~\cite{chance_constraints_pavone}. Finally, control inputs \( \boldsymbol{u}(k) \) are constrained to remain within specified limits \( [\boldsymbol{u}_{\text{min}}, \boldsymbol{u}_{\text{max}}] \). As we show in the next section, this formulation can operate effectively in various navigation tasks, including path tracking and obstacle avoidance.

\section{Chance-Constrained MPPI} \label{sec:chance_constrained_mppi}
This section outlines our GP-adapted MPPI framework, focusing on task-specific cost functions, uncertainty propagation, and chance constraints essential for effective path tracking and dynamic obstacle avoidance at high speeds.

\subsection{Robot Dynamics}
We design our control policy as an open-loop approach, where the control actions \( \boldsymbol{u}(k) \) are directly determined by MPPI optimization. This method simplifies the formulation by avoiding state-dependent control actions, ensuring that control inputs remain deterministic. We propagate the mean and covariance of robot states using a first-order Taylor expansion around the current system mean, similar to an extended Kalman filter update~\cite{gp_melanie}:
\begin{align} \label{eq:state_update}
    \boldsymbol{\mu}_x(k+1) &= \boldsymbol{f}(\boldsymbol{\mu}_x(k), \boldsymbol{u}(k)) + \overline{\boldsymbol{m}}(k) \\
    \boldsymbol{\Sigma}_x(k+1) &= \nabla \boldsymbol{f} \boldsymbol{\Sigma}_x(k) \nabla \boldsymbol{f}^\top + \text{diag}(0, 0, 0, \text{diag}(\overline{\boldsymbol{C}}(k))) \notag
\end{align}
Here, the first \(\text{diag}(\cdot)\) creates a diagonal matrix, while the second extracts diagonal elements from the GP covariance matrix \(\overline{\boldsymbol{C}}(k)\). The term \(\nabla \boldsymbol{f}\) is the gradient of the nominal system dynamics \(\boldsymbol{f}\) with respect to the mean robot state \(\boldsymbol{\mu}_x(k)\). We opt for a first-order Taylor expansion due to its computational efficiency, though more accurate methods like unscented transforms can also be used~\cite{gp_guided_mppi}.

\subsection{Navigation Cost Functions}\label{sec:navigation_cost_functions}
A key element of a custom cost function in our GP-MPPI framework for path tracking and obstacle avoidance is to penalize high GP variance. This encourages the controller to prioritize trajectories with lower uncertainty, promoting safer navigation where the model is more confident~\cite{nakka}. 

The tracking cost function \(c_{\text{track}}\) is defined as a sum of multiple components, each weighted by $\alpha\in[0,1]$ terms:
\begin{equation}
\begin{aligned}
c_{\text{track}} &= \alpha_0\, \text{Variance} + \alpha_1\, \text{Deviation} + \alpha_2\, \text{Slip} \\
                   &\ + \alpha_3\, \text{Safety} + \alpha_4\, \text{Speed} \\
                   &= \ \alpha_0 \, \text{trace}(\overline{\boldsymbol{C}}) + \alpha_1 \, d(\mu_x^X, \mu_x^Y) + \alpha_2 \, \sigma \\
                   &\ + \alpha_3 \, 0.9^k \, \mathbb{M}(\mu_x^X, \mu_x^Y) + \alpha_4 \, (v_{\text{desired}} - v_{\text{sampled}}) 
\end{aligned}
\label{eq:ctrack}
\end{equation}
Here, \(d(\mu_x^X, \mu_x^Y)\) is the normalized distance, where 0 refers to the lane center, 1 to boundaries, and values in-between reflect positions within the lane. This term penalizes deviations from the center, keeping the robot within the lane and away from boundaries. The slip \(\sigma\), defined as the ratio of lateral to longitudinal velocity, indicates the extent of sideways movement relative to forward motion. This cost penalizes high slip values, which suggest instability or loss of traction. \(\mathbb{M}(\mu_x^X, \mu_x^Y)\) is an indicator function that equals 1 when the robot crosses lane boundaries and 0 when it stays inside. The exponential term heavily penalizes trajectories that immediately exit the lane, while later violations are penalized less, allowing flexibility in path planning without excessively penalizing every lane departure. Lastly, the \((v_{\text{desired}} - v_{\text{sampled}})\) term penalizes deviations from the desired speed, encouraging the robot to maintain the target speed.

Similarly, the obstacle avoidance cost function \(c_{\text{avoidance}}\) is defined as a sum weighted by parameters $\beta \in [0,1]$:
\begin{equation}
\begin{aligned}
c_{\text{avoidance}} &= \beta_0\, \text{Variance} + \beta_1\, \text{Obstacle} \\
                       &\ + \beta_2\, \text{Stage} + \beta_3\, \text{Terminal} \\
                       &= \ \beta_0 \, \text{trace}(\overline{\boldsymbol{C}}) + \beta_1 \, \mathbb{I}(\mu_x^X, \mu_x^Y) \\
                       &\ + \beta_2 \, g(\mu_x^X, \mu_x^Y) + \beta_3 \, F(\mu_x^X, \mu_x^Y) 
\end{aligned}
\label{eq:cavoidance}
\end{equation}
Here, \(\mathbb{I}(\mu_x^X, \mu_x^Y)\) is 1 if the robot collides with an obstacle and 0 otherwise, thus imposing a high penalty for any trajectory that results in a collision. The function \(g(\mu_x^X, \mu_x^Y)\) penalizes distance-to-goal at each time step, encouraging the robot to move toward the target. Finally, \(F(\mu_x^X, \mu_x^Y)\) assigns a high terminal cost if the trajectory fails to reach the goal, and 0 if it succeeds. These task-specific cost functions guide the MPPI framework in selecting optimal control actions.

\subsection{Inequality Constraints Reformulation}
We now apply constraint tightening~\cite{smpc_survey} to reformulate safety constraints for path tracking and obstacle avoidance. These constraints are evaluated at each time step, but the time index is omitted for notational simplicity.
\begin{figure}[t]
    \centering
    \includegraphics[width=0.9\columnwidth]{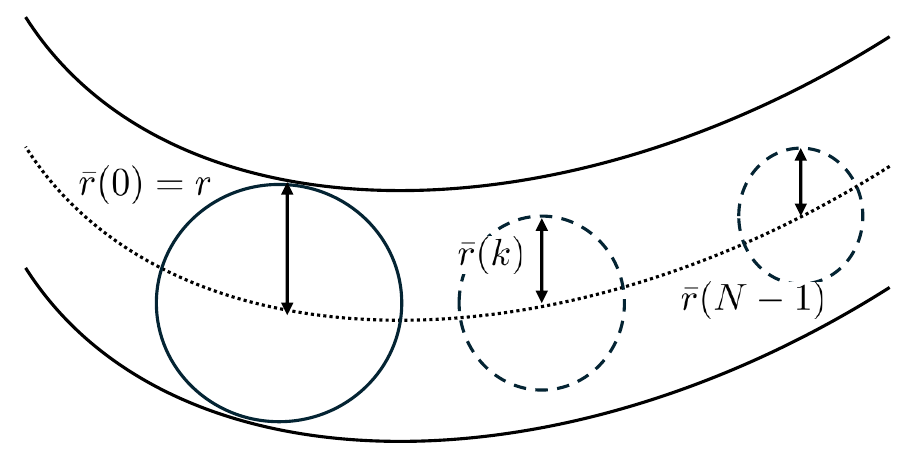} 
    \caption{Visualization of the track half-width reduction along the MPC horizon as a function of propagated variance.}
    \label{fig:track_shrinking}
    \vspace{-1.9mm}
\end{figure}

\textbf{Lane Boundary Constraint Reformulation:} The robot's ability to stay within the track is governed by the following inequality constraint:
\begin{equation}
\|(\mu_x^X, \mu_x^Y) - (X_c, Y_c)\|_2 \leq r
\end{equation}
where the 2-norm of the difference between the robot's instantaneous position \((\mu_x^X, \mu_x^Y)\) and the centerline coordinates \((X_c, Y_c)\) must remain less than the track's half-radius \(r\). Following the procedure in~\cite{path_tracking_tightening} for such ellipsoidal constraints, constraint tightening modifies the original constraint by replacing the half-radius \(r\) with a tightened \(\bar{r}\), where:
\begin{equation}
\bar{r} = r - \sqrt{\chi^2_2(p_x) \cdot \lambda_{\text{max}}(\boldsymbol{\Sigma}_{XY})}
\label{eq:lane_violation_tightening}
\end{equation}
Here, \(\chi^2_2(p_x)\) is the quantile function of the chi-squared distribution with two degrees of freedom and \(\lambda_{\text{max}}(\boldsymbol{\Sigma}_{XY})\) is the maximum eigenvalue of the state covariance matrix \(\boldsymbol{\Sigma}_{XY}\), which reflects the uncertainty in the robot's predicted \(XY\)-position. This approach ensures that the safety constraints are met with the probability threshold \(p_x\) specified in Eq.~\ref{eq:generic_chance_constraint}. Given the modified track radius \(\bar{r}\), we now adapt the corresponding safety term in Eq.~\ref{eq:ctrack} by injecting this new radius. This guarantees that the safety cost obeys the tightened constraint, penalizing any deviation from the modified safety boundary. Fig.~\ref{fig:track_shrinking} visually depicts this process.

\begin{figure}[t]
    \centering
\includegraphics[width=0.6\linewidth]{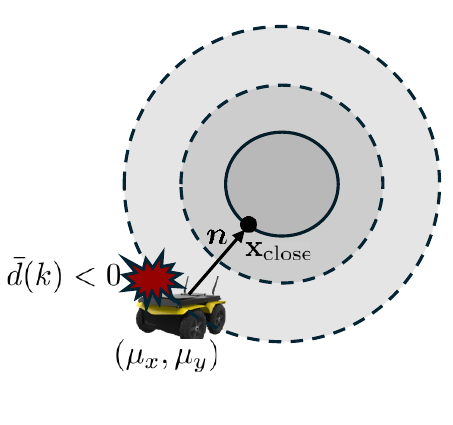} 
    \caption{Visualization of the modified signed distance function, leading to a predicted collision at time step \( k \).}
    \label{fig:obstacle_inflation}
    \vspace{-3.8mm}
\end{figure}

\textbf{Obstacle Avoidance Reformulation:} We follow a similar procedure to~\cite{chance_constraints_pavone} to modify the safety buffer for obstacle avoidance. The signed distance function $d$ represents the shortest distance between the robot and an obstacle, with a negative value indicating a collision.  The closest point on the obstacle boundary is determined, \(\mathbf{x}_{\text{close}} = (X_{\text{close}}, Y_{\text{close}})\), and we define a unit vector \(\mathbf{n}\) pointing from the robot toward this obstacle:
\begin{equation}
\label{eq:obs_avoidance_tightening}
\mathbf{n} = \frac{(\mu_x - X_{\text{close}}, \mu_y - Y_{\text{close}})}{d}
\end{equation}
Finding the closest point \(\mathbf{x}_{\text{close}}\) and the distance \(d\) can be non-trivial for arbitrary shapes. We simplify this calculation by approximating obstacles as circles. To account for uncertainty in the robot's location, we modify the obstacle avoidance constraint as follows:
\begin{equation}
\label{eq:new_obs_cons}
\bar{d} = d - \Phi^{-1}(p_x) \cdot \sqrt{\mathbf{n}^\top \boldsymbol{\Sigma}_{XY} \mathbf{n}} > 0
\end{equation}
where \(\Phi^{-1}(p_x)\) is the inverse cumulative distribution function associated with the probability threshold \(p_x\), and the term \(\sqrt{\mathbf{n}^\top \boldsymbol{\Sigma}_{XY} \mathbf{n}}\) accounts for uncertainty in the robot's predicted position. Hence, Eq.~\ref{eq:new_obs_cons} represents the modified signed distance function to maintain a collision-free state. As a result, we update the obstacle avoidance check in Eq.~\ref{eq:cavoidance} based on this newly calculated threshold. Fig.~\ref{fig:obstacle_inflation} further clarifies this process.

\subsection{Overall Stochastic MPPI Algorithm}
The key steps of the Stochastic MPPI algorithm to solve the navigation problem outlined in Eqs.~\ref{eq:cost_function} to~\ref{eq:initial_condition} are summarized in Algorithm~\ref{alg:stochastic_mppi}. Our sampling based motion planner was implemented on an NVIDIA RTX 3080 GPU, using the Just-In-Time compiler, Numba, to develop CUDA kernels, significantly speeding up computation~\cite{evora,numba}. The parallelized components include the sampling of MPPI trajectories, cost evaluation for each trajectory, and GP inference for each terrain, all of which were executed on the GPU. Constraint tightening is based on the previous iteration's optimal control sequence, allowing pre-computed safety adjustments to be applied during the current iteration's optimization~\cite{gp_melanie}. Depending on the task--path tracking or obstacle avoidance--the appropriate cost function and constraint tightening method are applied. Certain MPPI steps are briefly mentioned but not elaborated here due to space constraints; for these, we refer the interested reader to~\cite{mppi_basic}.

\begin{algorithm}[t]
\caption{Stochastic MPPI}
\label{alg:stochastic_mppi}
\begin{algorithmic}
\Require Initial Control Sequence: \( \boldsymbol{U} = \{\boldsymbol{u}(0), \ldots, \boldsymbol{u}(N-1)\} \)
\Require Parameters: \( \boldsymbol{\alpha} \), \( \boldsymbol{\beta} \), \( \boldsymbol{\Sigma}_{\text{sim}} \), \( \lambda \), \(p_x\)
\Require Initial Terrain Weights: \( w_0, \ldots, w_M \), Eq.~\ref{eq:convex_opt}
\While{goal not reached}
    \State \texttt{//Step 1: Main MPPI Loop}
    \State Get State Estimate: \( \boldsymbol{\mu}_x(0) = \boldsymbol{x}(0) \) and \( \boldsymbol{\Sigma}_x(0) = \boldsymbol{0} \)
    \For{$s \gets 0$ to Number of Samples, \( S \)}
        \State Sample \( \boldsymbol{\epsilon}^s = \{\boldsymbol{\epsilon}_0^s, \ldots, \boldsymbol{\epsilon}_{N-1}^s\}, \boldsymbol{\epsilon}_k^s \sim \mathcal{N}(0, \boldsymbol{\Sigma}_{\text{sim}}) \)
        \For{$k \gets 0$ to $N-1$}
            \State \( \boldsymbol{u}^s(k) = \boldsymbol{u}(k) + \boldsymbol{\epsilon}_k^s \)
            \State \( \overline{\boldsymbol{m}}^s(k), \overline{\boldsymbol{C}}^s(k) \gets \text{GP output, Eq.~\ref{eq:weighted_gp_eqns}} \)
            \State \( \boldsymbol{\mu}_x^s(k+1) \gets \text{Eq.~\ref{eq:state_update}} \)
        \EndFor
        \State Evaluate cost: \( c_{\text{track}}^s \) or \( c_{\text{avoidance}}^s \) Eqs.~\ref{eq:ctrack},~\ref{eq:cavoidance}
        \State Compute MPPI Trajectory Weights~\cite{mppi_basic}
    \EndFor
    \State \( \boldsymbol{U} \gets \text{Weighted sum of MPPI sample costs~\cite{mppi_basic}} \)


    \State \texttt{//Step 2: Constraint Tightening Loop}
    \For{$k \gets 0$ to $N-1$}
        \State \( \boldsymbol{\Sigma}_x(k+1) \gets \text{Eq.~\ref{eq:state_update}} \)
        \State Adjust Safety Constraints: Eq.~\ref{eq:lane_violation_tightening} or Eq.~\ref{eq:new_obs_cons}
        \State Modify \( c_{\text{track}} \) or \( c_{\text{avoidance}} \) for tightened constraints
    \EndFor
    \vspace{0.25em} 
    \State \texttt{//Step 3: Terrain Estimation}
    \State Optimize Terrain Weights: Eq.~\ref{eq:convex_opt}
\EndWhile
\end{algorithmic}
\end{algorithm}

\section{Experiments and Discussion}\label{sec:experiments}
In this section, we present both simulation and hardware experiments conducted to evaluate our GP-MPPI planner. Please refer to our supplementary video for additional results and demonstrations.

\subsection{Simulation Experiments}\label{sec:simulation_experiments}
We used a dataset from Rabiee et al.~\cite{rabiee}, containing 6 km of motion data for the Clearpath Jackal skid-steer robot across Tile, Asphalt, and Grass terrains. This dataset provided the commanded and ground truth velocities needed to train our GPR model.

We benchmarked our chance-constrained GP-MPPI against two kinematic models: the Extended Differential Drive with five parameters (EDD5) model and the standard unicycle model. The EDD5 model relates commanded velocties to actual wheel velocities using five parameters obtained through least squares regression. Both kinematic models were integrated into the MPPI planner for comparison. For the EDD5 model, we used 300 data points per terrain, totaling 900 data points across the three terrains.

For our GP-based model, we trained a batch GP using 300 shared data points across terrains, each a four-dimensional input of current linear velocity, current angular velocity, commanded linear velocity, and commanded angular velocity. The corresponding outputs, however, were terrain-specific, resulting in a $300\times6$ output matrix that captures the residual linear and angular velocity errors for each of the three terrains. This multi-output Batch GP model, trained using the GPyTorch library~\cite{gpytorch}, allowed the GP-MPPI planner to generalize across varying terrains while maintaining computational efficiency. To ensure fair comparison, the MPPI cost function parameters were kept identical for all models.

In addition to using GP and EDD5 models for motion prediction, we also trained a neural network to model the relationship between commanded and measured velocities. The motivation behind this was twofold. First, the neural network was trained on the full 6 km of motion data, resulting in more accurate predictions when evaluated on a test set. Second, this approach follows the methodology used in the original MPPI paper~\cite{mppi_basic}, where a neural network was employed for similar tasks. Therefore, we used this neural network for ground truth velocity estimation once the commanded velocities were chosen. 

We conducted two experiments: path tracking with square and circular tracks, as well as static obstacle avoidance. For the tracking experiments, commanded speeds were set to 2 m/s. Circular paths were selected to evaluate a planner's ability to maintain continuous motion at constant speeds, providing insight into terrain modeling accuracy. While square paths focus on handling sharp 90-degree turns, which is an important test-case for real-world navigation. For obstacle avoidance, the cost functions outlined in section~\ref{sec:navigation_cost_functions} encouraged the robot to minimize the time to reach the goal. 

\begin{figure}
    \centering
    \begin{subfigure}{0.45\linewidth}
        \includegraphics[width=\linewidth,height=4cm]{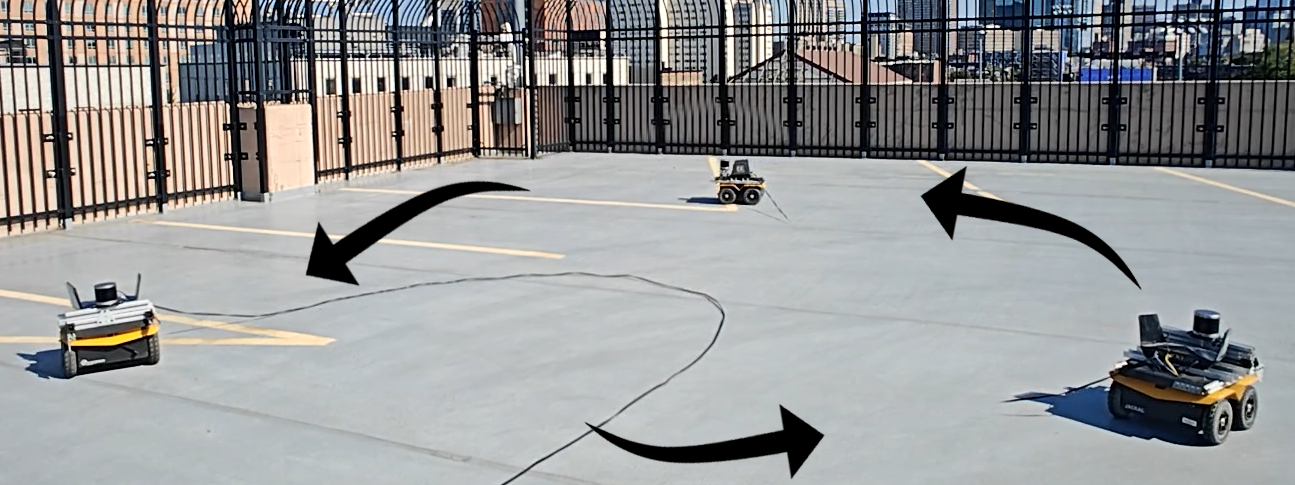}
        
        \label{fig:circular_path_tracking}
    \end{subfigure}
    \begin{subfigure}{0.45\linewidth}
        \includegraphics[width=\linewidth]{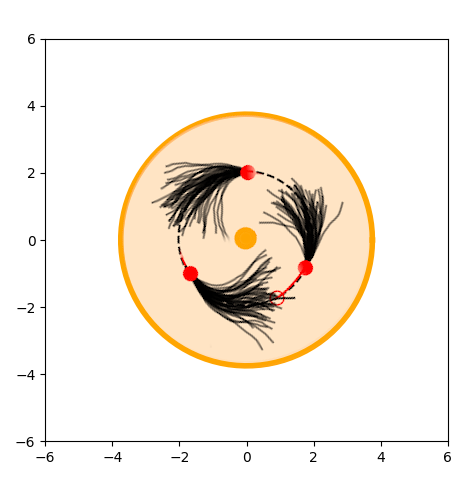}
        
        \label{fig:circular_path_tracking_gp_plot}
    \end{subfigure}
    \caption{Circular paths on tiles tracked by a skid-steer robot operating at a reference linear speed of 2 m/s (top speed of robot) and an angular speed of 1.25 rad/s.}
    \label{fig:tracking_hardware}
    \vspace{-2mm}
\end{figure}

Table 1 presents our simulation experiment's results. For path tracking, we conducted a 100 m trial across Tile, Asphalt, and Grass terrains, comparing performance of the three planners. We measured tracking accuracy as the RMSE error between the closest lane center and the actual robot position. For obstacle avoidance, the robot was tasked with navigating up to five randomly placed obstacles from start-to-goal across three terrains using the three planners. We conducted 100 trials, varying the number, shape, size, and location of obstacles between the robot and goal. We evaluated time-to-goal and the number of collisions for each trial. Across both path tracking and obstacle avoidance tasks, GP-MPPI consistently demonstrated superior performance, as indicated by lower RMSE values, fewer collisions, and lower time-to-goal. These findings highlight our GP-MPPI method's adaptability to diverse terrains and complex scenarios.

\subsection{Hardware Validation}\label{sec:hardware_validation}

\begin{table*}[t]
\caption{Simulation Results: RMSE, Success Rate, and Time-to-Goal.}
\centering
\begin{tabularx}{0.8\paperwidth}{|c|m{2em}|m{2em}|m{2em}|m{2em}|m{2em}|m{2em}|m{2em}|m{2em}|m{2em}|m{3.5em}|m{3.5em}|m{3.5em}|}
    \hline
    \multirow{2}{*}{} & \multicolumn{3}{|X|}{\textbf{Circle Track (100m)}} 
    & \multicolumn{3}{|X|}{\textbf{Square Track (100m)}} 
    & \multicolumn{6}{|c|}{\textbf{Obstacle Avoidance (100 trials)}} \\
    \hline

    & \multicolumn{3}{|c|}{\textbf{RMSE (mm)}}
    & \multicolumn{3}{|c|}{\textbf{RMSE (mm)}}
    & \multicolumn{3}{|c|}{\textbf{Success Rate}}
    & \multicolumn{3}{|c|}{\textbf{(Success) Time-to-Goal (s)}}\\ \hline
    
    \textbf{Terrain} & \textbf{EDD} & \textbf{UNI} & \textbf{GP} & \textbf{EDD} & \textbf{UNI} & \textbf{GP} & \textbf{EDD} & \textbf{UNI} & \textbf{GP} & \textbf{EDD} & \textbf{UNI} & \textbf{GP} \\
    \hline

    \text{Grass} & 92 & 113 & \textbf{31} & 89 & 80 & \textbf{39} & 25/33 & 22/33 & \textbf{29/33} &  4.91$\pm$0.2 & 5.01$\pm{}$0.1 & \textbf{4.89$\pm$0.2} \\
    \hline

    \text{Tile} & 73 & 94 & \textbf{24} & 31 & 22 & \textbf{10} & 29/33 & 25/33 & \textbf{33/33} & 5.21$\pm$0.1 & \textbf{5.03$\pm$0.2} & 5.23$\pm$0.1 \\
    \hline

    \text{Asphalt} & 72 & 73 & \textbf{26} & 62 & 30 & \textbf{12} & 31/34 & 27/34 & \textbf{32/34} & 5.19$\pm$0.2 & 5.18$\pm$0.1 & \textbf{5.12$\pm$0.2} \\
    \hline
\end{tabularx}
\end{table*}

\begin{figure*}[t]
    \centering
    \begin{subfigure}{0.26\linewidth}
        \includegraphics[width=\linewidth,height=4.cm]{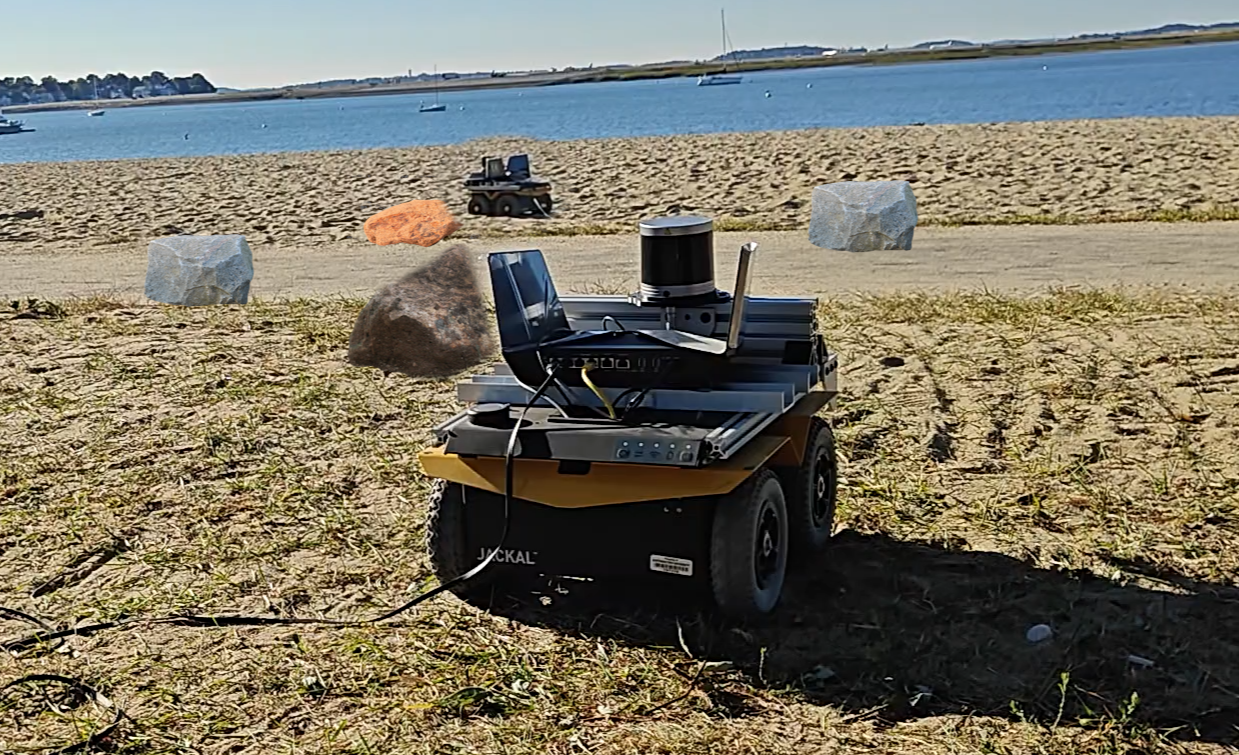}
        \caption{}
        \label{fig:beach_visuals}
    \end{subfigure}
    \begin{subfigure}{0.25\linewidth}
        \includegraphics[width=\linewidth,height=4cm]{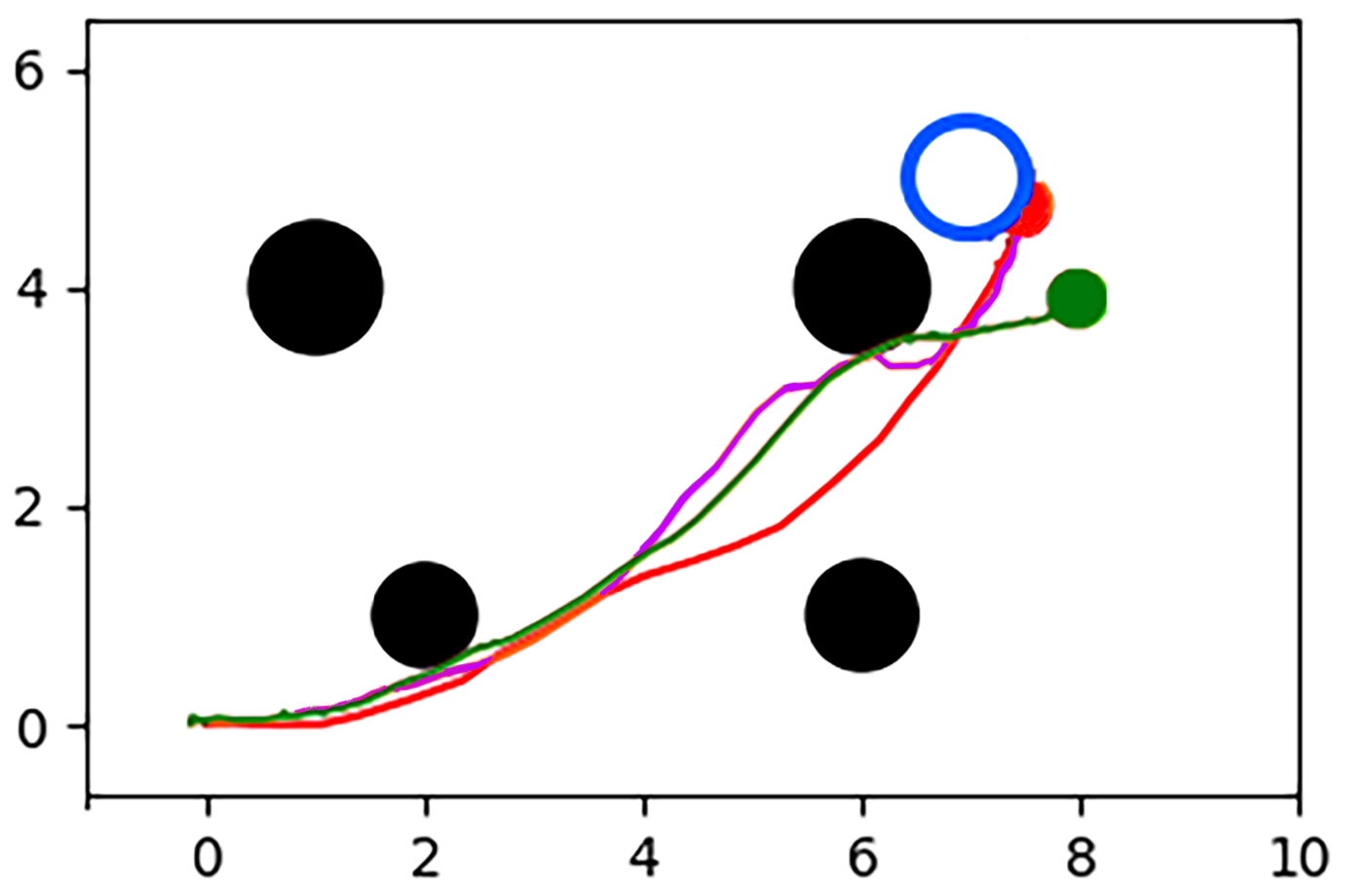}
        \caption{}
        \label{fig:beach_plots}
    \end{subfigure}
    \begin{subfigure}{0.4\linewidth}
        \includegraphics[width=\linewidth,height=4cm]{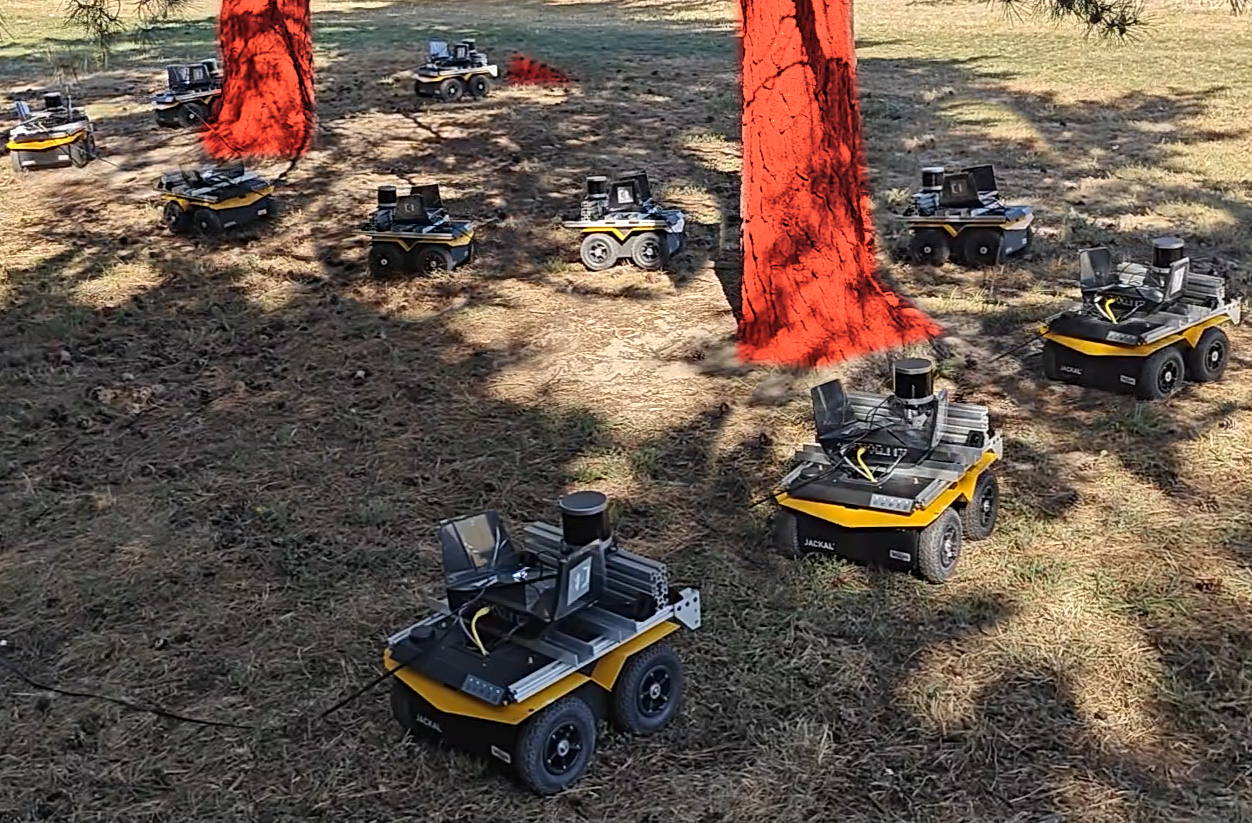}
        \caption{}
        \label{fig:slaloming}
    \end{subfigure}
    \caption{ (a) Navigating a beach terrain around virtual obstacles highlighted as stones. (b) Trajectories from all planners: Unicycle-MPPI (green) collides with obstacles, EDD5-MPPI (pink) takes longer and passes close to obstacles, while GP-MPPI (red) avoids obstacles and reaches the goal faster. (c) GP-MPPI based slaloming around trees on a bushy terrain.}
    \label{fig:main_hardware}
    \vspace{-1.1mm}
\end{figure*}

We deployed our chance-constrained GP-MPPI algorithm on a Clearpath Jackal skid-steer robot equipped with a Robosense 32 channel LiDAR. Ground truth positions and velocities were estimated using the direct LiDAR odometry package~\cite{direct_lidar_odometry}. The training setup for the EDD5 and GP models, including the number of training points, GP library, and GPU-based parallel processing, was identical to the simulation setup. For circular path tracking experiments, as shown in Fig.~\ref{fig:tracking_hardware}, the robot attained a tracking RMSE of 55 mm, closely matching the results observed in simulation.

In the obstacle avoidance experiments, obstacles were modeled as virtual objects to avoid the need for real-time obstacle detection and to minimize the risk of robot damage during trials. These experiments were conducted across multiple terrains, including grass, paved roads, and dense sand, as the robot navigated from a starting point to a goal. The sharp turn through dense sand, in particular, emphasized the importance of live terrain estimation, as the skid properties of sand were not explicitly modeled. Despite this challenge, the robot successfully navigated the surface, demonstrating the versatility of our approach. The experimental setup is shown in Fig.~\ref{fig:beach_visuals}, with the resulting trajectories from different planners illustrated in Fig.~\ref{fig:beach_plots}.

In these trials, the Unicycle-MPPI planner had the worst performance, colliding with obstacles (visualized as stones in Fig.~\ref{fig:beach_visuals}). The EDD5-MPPI planner avoided collisions but approached obstacles too closely, reaching the goal in 9.24 seconds. In contrast, the GP-MPPI planner maintained a safe distance from obstacles and reached the goal in 6.5 seconds at an average speed of 1.76 m/s. For obstacle avoidance on a bushy surface with real trees, as shown in Fig.~\ref{fig:slaloming}, we limited trials to the GP-MPPI algorithm due to concerns over potential collisions and damage when using other methods. The average speed for this experiment was recorded to be 1.85 m/s. These experiments demonstrate the robustness of GP-MPPI in handling challenging, real-world environments.

\subsection{Discussion}
In this section, we discuss why GP-MPPI outperformed other methods in our experiments on path tracking and collision avoidance. First, GPs accurately model complex tire-terrain interactions that simpler models, like EDD5 and unicycle, cannot. Additionally, GPs account for uncertainty by predicting higher variance estimates in areas where the model is less confident, prompting chance constraints in our stochastic MPPI formulation to adjust safety buffers by either reducing track dimensions or increasing distance-to-obstacle thresholds. Unlike other data-driven methods, such as MPPI with neural networks, GP-MPPI balances safety and performance autonomously based on state uncertainty and a probability threshold \(p_x\). Finally, GPU-based parallelization ensures real-time performance, efficiently handling the non-convex optimization problem.

\section{Conclusion}\label{sec:conclusion}
This paper proposes a chance-constrained GP-MPPI algorithm for high-speed multi-terrain skid-steer robot navigation. Our approach was benchmarked against state-of-the-art kinematic model-based MPPI methods, demonstrating improved real-time performance in both path tracking and obstacle avoidance. Future work could extend this GP-MPPI algorithm to other platforms, such as quadrotors and race cars, enabling broader applications of stochastic motion planning. Furthermore, the flexibility of MPPI in incorporating arbitrary cost functions could offer ways of integrating visual perception data into the framework for context-aware decision-making. One limitation identified during hardware testing was that the robot platform struggled to handle the noisy MPPI outputs~\cite{mppi_basic} required for highly dynamic 90-degree turns on a square path. To address this, we plan to extend our approach using recent MPPI variants that inherently generate smoother motion~\cite{smooth_mppi}.

\bibliographystyle{IEEEtran}
\bibliography{IEEEabrv,references}

\end{document}